\documentclass[letterpaper, 10 pt, conference]{ieeeconf}  

\IEEEoverridecommandlockouts  
\usepackage{amsmath}
\usepackage{bm}
\usepackage{amsfonts}
\usepackage{amssymb}
\usepackage[nospace]{cite}
\usepackage{url}
\usepackage{xcolor}
\usepackage{fancyhdr}
\usepackage[hidelinks]{hyperref}

\usepackage{adjustbox}
\usepackage{graphicx}
\usepackage{tensor}
\usepackage{tabularx}

\usepackage[utf8]{inputenc}
\usepackage[OT1]{fontenc}
\usepackage[final]{microtype}
\usepackage[nolist,nohyperlinks]{acronym}
\acrodef{vae}[VAE]{Variational Autoencoder}
\acrodef{sevae}[seVAE]{semantically-enhanced Variational Autoencoder}
\acrodef{cpn}[CPN]{Collision Prediction Network}
\usepackage{multirow}
\usepackage{makecell}

\usepackage{booktabs}

\usepackage[binary-units]{siunitx}

\usepackage{lipsum}

\DeclareMathAlphabet{\pazocal}{OMS}{zplm}{m}{n}

\newcommand{\Bs}{\pazocal{B}}

\newcommand{\Vs}{\pazocal{V}}

\title{\LARGE \bf
Semantically-enhanced Deep Collision Prediction for Autonomous Navigation using Aerial Robots
}

\author{Mihir Kulkarni, Huan Nguyen, and Kostas Alexis 
\thanks{This material was supported by a) the AFOSR Award No. FA8655-21-1-7033 and b) the Research Council of Norway Award NO-321435.}
\thanks{The authors are with the Autonomous Robots Lab, Norwegian University of Science and Technology (NTNU), O. S. Bragstads Plass 2D, 7034, Trondheim, Norway {\tt\small mihir.kulkarni@ntnu.no}}
}

\begin{document}

\maketitle
\thispagestyle{empty}
\pagestyle{empty}

\begin{abstract}
This paper contributes a novel and modularized learning-based method for aerial robots navigating cluttered environments containing hard-to-perceive thin obstacles without assuming access to a map or the full pose estimation of the robot. The proposed solution builds upon a semantically-enhanced Variational Autoencoder that is trained with both real-world and simulated depth images to compress the input data, while preserving semantically-labeled thin obstacles and handling invalid pixels in the depth sensor's output. This compressed representation, in addition to the robot's partial state involving its linear/angular velocities and its attitude are then utilized to train an uncertainty-aware $3$D Collision Prediction Network in simulation to predict collision scores for candidate action sequences in a predefined motion primitives library. A set of simulation and experimental studies in cluttered environments with various sizes and types of obstacles, including multiple hard-to-perceive thin objects, were conducted to evaluate the performance of the proposed method and compare against an end-to-end trained baseline. The results demonstrate the benefits of the proposed semantically-enhanced deep collision prediction for learning-based autonomous navigation.

\end{abstract}

\section{Introduction}\label{sec:intro}
The rapid progress in the field of aerial robotics has enabled their successful utilization in diverse applications including search-and-rescue~\cite{delmerico2019jfr,tranzatto2022cerberus}, inspection~\cite{SIP_AURO_2015,caprari2012highly}, and forest monitoring~\cite{Zhang2016Forest,aucone2023dronedna}. Keys to this success have been techniques to plan paths, often via sampling-based methods and motion primitives~\cite{RRT,karaman2011rrt_star,paranjape2015motion}, within dense $3\textrm{D}$ maps~\cite{voxblox, hornung13auro, museth2013vdb,florence2018nanomap}. Yet, despite the progress, most approaches are limited by a) the need for consistent localization and mapping, b) the latency such systems introduce, as well as c) the maximum resolution of the map used for collision checking and the computational burden of increasing such resolution in order for narrow cross-section (thin) obstacles (e.g., wires, railings, rods, tree branches) to be represented. Recently, we have seen great advances coming from the field of deep learning for collision-free flight tailored to a host of demanding tasks including drone racing~\cite{Foehn2021AlphaDrone}. Yet, most such methods also assume consistent pose estimation~\cite{Loquercio2021Science}, and often rely fully on simulated data for training~\cite{ORACLE_ICRA2022} which, however, may fail to capture significant imperfections of range sensing such as holes in depth maps~\cite{yang2018segstereo,ahn2019analysis}, while there is a lack of focus on methods that explicitly tackle avoiding thin, hard-to-perceive, obstacles.

\begin{figure}[t!]
\includegraphics[width=\columnwidth]{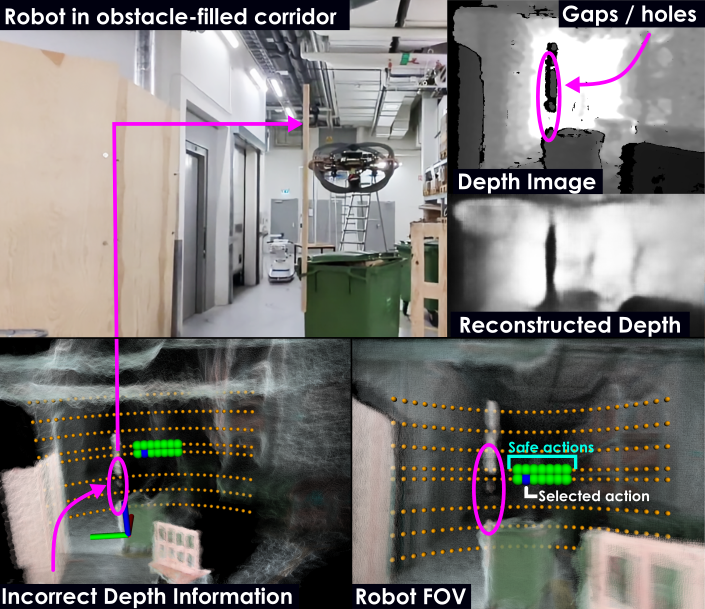}
\vspace{-5ex}
\caption{Instance of an experiment demonstrating collision-free flight in a cluttered environment further involving thin obstacles (e.g., vertical wooden sticks) with the robot only exploiting real-time depth maps and estimates of its linear and angular velocities, as well as attitude.}
\label{fig:intro}
\vspace{-6ex}
\end{figure}


Motivated by the above, this work contributes a modularized learning-based navigation method that a) is capable of navigating cluttered environments involving hard-to-perceive thin obstacles, while b) not requiring access to a (online or offline) map of the environment or the estimation of the full pose of the robot but instead relies solely on real-time depth observations and a partial estimate of the robot's linear/angular velocities and attitude. While the community has considered the explicit detection of certain thin obstacles (e.g., wires are studied in~\cite{Madaan2017WireDU,Kulkarni2020} using RGB images), the challenge of safe navigation in cluttered scenes with complex topologies and morphologies of thin obstacles is not solved. Furthermore, for such structures, acquiring accurate depth maps using real sensors is particularly difficult due to their narrow cross-sections. To overcome this limitation and considering the use of depth cameras, we first create and aggregate both real and simulated datasets of depth frames with pixel-wise labels on thin obstacles. Then, we contribute a semantically-enhanced Variational Autoencoder (seVAE) that a) learns to compress, encode and reconstruct depth maps while preserving thin obstacle features, b) utilizes both real and simulated depth data for training along with pixel-wise labels reflecting thin obstacles (when available), and c) can work directly with raw depth frames while remaining robust to complex cases of sensor noise and errors including holes in the depth maps. Through the encoder of the seVAE, the method arrives --at inference time-- to a low-dimensionality latent space that encodes the information for collision avoidance including against thin obstacles. This latent space is then combined with a Collision Prediction Network (CPN) building upon our previous work called ``ORACLE''~\cite{ORACLE_ICRA2022}. The CPN utilizes the compressed latent space of the encoder combined with the robot's partial state (considering its uncertainty) and provides collision scores for candidate $3\textrm{D}$ action sequences from a motion library thus enabling safe autonomous flight.

To validate the proposed approach, a set of evaluation studies, both in simulation and experimentally, were conducted. In simulation, the method was tested in environments with increasing density of obstacles and its results are compared against an end-to-end trained $3\textrm{D}$ extension of ORACLE (using simulated depth images only). Two real-world experiments were also performed to test the new method in densely cluttered environments with various sizes and types of obstacles. Similarly, the performance of the new method in these experiments is compared against that of the end-to-end trained ORACLE demonstrating the value added by the ability to incorporate real data in the training phase to enhance robustness against sensor imperfections.

In the remainder of this paper, Section~\ref{sec:related} presents related work, followed by the proposed modularized learning-based navigation solution in Section~\ref{sec:approach}. Evaluation studies are detailed in Section~\ref{sec:evaluation}, and conclusions in Section~\ref{sec:conclusions}.


\section{Related Work}\label{sec:related}
A set of contributions in learning-based navigation relate to this work, alongside efforts to enable avoidance of thin obstacles. Many works have employed deep learning techniques to tackle the issue of autonomous robot navigation, with a particular focus on using RGB/depth cameras. These sensors have garnered increased attention due to their low cost, low power consumption, and lightweight design.
The authors in~\cite{Loquercio2021Science,Tolani2021visual} use imitation learning to generate collision-free smooth trajectories from realistic simulated data.
Though much effort has been devoted to obtaining more realistic simulation images~\cite{Zakharov2022photo_realistic,Bousmalis2018DomainAdaptation}, there persist effects that are difficult to model such as missing depth pixels when observing thin obstacles or low-texture, shiny surfaces.
In~\cite{Fereshteh2017CAD2RL}, reactive navigation policies are learned through reinforcement learning and domain randomization using RGB image inputs.
On the other hand,~\cite{Gandhi2017learning2fly,Kahn2021BADGR,Kahn2021LAND} predict probabilities of events such as collisions, going over bumpy terrain, or human disengagement from real-world RGB data.
Our approach also leverages real depth data in the training pipeline, however, part of the depth frames and the data collection where the robot dynamics are rolled out and collision events happen is performed in simulation.
The authors in~\cite{Hoeller2021representation} propose another method to mix real and simulated depth images for learning an environment representation for legged robot navigation. In this work, we focus on using both the real and simulated depth images augmented with the labeled semantic masks to ensure the reconstruction of hard-to-perceive obstacles such as thin objects in the environment representation.

Of further relevance to this work are also contributions proposing solutions to allow robots to detect and avoid thin obstacles in the environments. The authors in~\cite{Zhou2017FastAT} present a thin-structure obstacle detection and 3D reconstruction method based on an edge detector and edge-based visual odometry techniques. On the other hand, a multi-view algorithm for wire reconstruction using a parametric catenary model is illustrated in~\cite{Madaan2019MultiviewRO}. The work in~\cite{Landry2016MIP} proposes a planning approach using mixed integer programming to derive collision-free trajectories for a quadrotor to fly in environments populated with thin strings, however, the locations of the obstacles are given beforehand as convex hulls. The authors in~\cite{Dubey2018droan} use a monocular wire detection method based on synthetic data and a dilated convolutional neural network (CNN)~\cite{Madaan2017WireDU} to perceive thin-wire obstacles. The obstacles are then represented by a disparity-space representation for collision checking. A separate planning module evaluates a pre-computed trajectory library to choose and execute the best trajectory.


\section{Proposed Approach}\label{sec:approach}
Our proposed modularized solution for learning-based navigation builds upon a \ac{vae}~\cite{doersch2016tutorial,kingma2013auto} module that can be independently trained using both real and simulated data to produce a compressed latent representation of a raw depth image that is then exploited for collision prediction. Semantic labels can also be added to the images --when available-- to ensure reconstruction of hard-to-perceive thin obstacles (e.g., skinny rods and wires~\cite{Kulkarni2020}). This representation is able to reconstruct complex images without missing out on narrow cross-section (thin) obstacles, while being robust to sensor noise that arises from the sensor's inability to reconstruct depth for textureless or reflective regions and stereo shadows. This latent representation is then used to train a $3$D \ac{cpn} in simulation to predict collision scores for candidate action sequences based on a library of motion primitives. Building upon our prior work called ORACLE~\cite{ORACLE_ICRA2022}, the method does not require the full robot state or an environment map but only real-time depth frames and access to an estimate of the partial robot state involving the system's attitude, as well as angular and linear velocities. Simultaneously, it takes into account the uncertainty associated with this partial state when predicting collision scores. Given the above and a goal vector provided by a high-level planner~\cite{GBPLANNER_JFR_2020,achtelik2014motion} or by an operator, the optimal collision-free action sequence is selected and executed by the robot in a receding horizon manner.

\subsection{Semantically-enhanced Variational Autoencoder}\label{sec:vae}
\ac{vae}s are powerful tools that can allow encoding high-dimensional input data in a compressed representation. In this work, we show that a complex depth image with a typical resolution (of $270\times 480$ pixels) can be sufficiently represented by a highly compressed latent representation (here $128$ variables) while preserving features from hard-to-perceive thin obstacles. This latent representation can subsequently be used for predicting collisions given a set of candidate action sequences. We develop an approach to combine data from both simulated and real sensor observations, allowing our method to be a) robust to sensor noise, and at the same time b) able to utilize additional information such as instance semantic labels --primarily from simulators-- thus being capable of reconstructing thin objects. In contrast, non-modularized methods that train a collision predictor end-to-end such as~\cite{ORACLE_ICRA2022} may not allow to explicitly promote a focus on such semantically-driven thin obstacles or conveniently utilize real sensor data and their imperfections.

\begin{figure*}[ht]
\includegraphics[width=\textwidth]{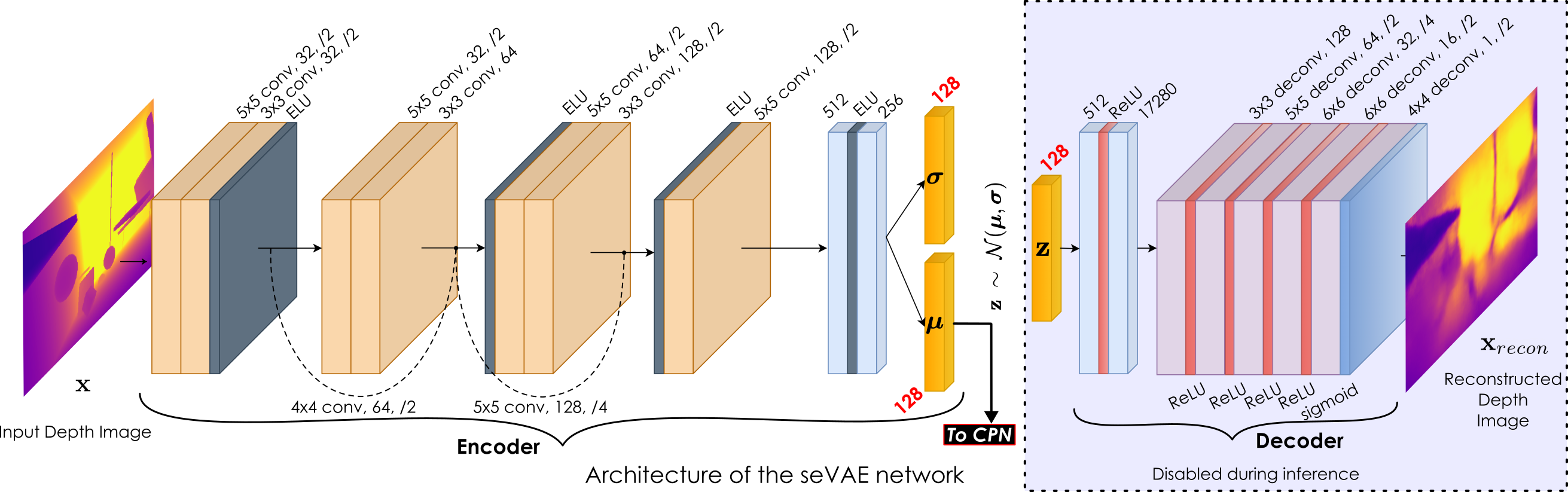}
\vspace{-4ex}
\caption{Proposed architecture for the \ac{sevae} for collision prediction. The convolutional and deconvolutional layers' hyperparameters are represented in the format ($\mathfrak{a}$ $\times$ $\mathfrak{b}$ (de)conv$\mathfrak{, c, /d}$), where $\mathfrak{a}$ $\times$ $\mathfrak{b}$ refers to the kernel size, $\mathfrak{c}$ refers to the number of channels, and $\mathfrak{d}$ refers to the stride length. The dense layers only have the layer size mentioned alongside. A latent space $\mathbf{z}$ of $128$ dimensions is sampled from the distribution with mean $\bm{\mu}$ and std. dev. $\bm{\sigma}$ and used to reconstruct the image by the decoder.}
\vspace{-4ex}
\label{fig:VAE_network_architecture}
\end{figure*}

Representing a compressed latent space starting from high-dimensional depth data using hand-tuned parameters is difficult, given that the distribution generating the dataset of depth images is practically intractable to compute. Thus, we utilize a \ac{vae} to learn a latent representation that can be used for image reconstruction. We consider a dataset $\mathbf{X} = \{\mathbf{x}^{(i)}, \mathbf{x}_{val}^{(i)}, \mathbf{x}_{seg}^{(i)}\}^N_{i=1}$ consisting of samples of discrete images $\mathbf{x}^{(i)}$ of dimensions $H\times W$ (here $270\times480$ pixels), the mask of valid pixels from the input data $\mathbf{x}_{val}^{(i)}$, and the instance segmentation mask for each thin obstacle instance $\mathbf{x}_{seg}^{(i)}$ (when available). The pixels with defined depth information from the sensor are referred to as valid pixels and the pixels with missing depth values are referred to as invalid pixels. We assume that this dataset is generated by some random process involving an unobserved random variable $\mathbf{z}$ with $J$ dimensions (here $J=128$). We model this variable $\mathbf{z}$ as the latent code for our representation. We assume that a depth image $\mathbf{x}$ is generated from $\mathbf{z}$ by some generative model. A probabilistic decoder $p_{\bm{\theta}}(\mathbf{x}|\mathbf{z})$ produces a distribution over the possible values of $\mathbf{x}$. Since the true value of the posterior $p_{\bm{\theta}}(\mathbf{z}|\mathbf{x})$ is unknown, we assume an approximate posterior $q_{\bm{\phi}}(\mathbf{z}| \mathbf{x})$, to be a multivariate Gaussian with a diagonal covariance as:

\small
\begin{equation}
    \log q_{\bm{\phi}}(\mathbf{z}| \mathbf{x}^{(i)}) = \log \mathcal{N}(\mathbf{z}; \bm{\mu}^{(i)}, {\bm{\sigma}}^{2(i)}\mathbf{I}),
\end{equation}
\normalsize
where the mean and the standard deviation of the approximate posterior, $\bm{\mu}$ and $\bm{\sigma}$, are outputs of the encoding neural network for the image $\mathbf{x}$ with parameters $\bm{\phi}$. In this realization, we use the reparameterization trick to obtain the values of $\bm{\mu}^{(i)}$ and $\log({\bm{\sigma}}^{2(i)})$ from the network~\cite{doersch2016tutorial}. We then sample $\mathbf{z}^{(i)} \sim q_{\bm{\phi}}(\mathbf{z} | \mathbf{x}^{(i)})$ using $\mathbf{z}^{(i)} = \bm{\mu}^{(i)} + \bm{\sigma}^{(i)}\odot\bm{\epsilon}$, ($\odot$ is the element-wise multiplication operator), with an auxiliary noise variable $\bm{\epsilon} \sim \mathcal{N}(\mathbf{0},\mathbf{I})$.  A suitable encoder for this architecture is designed by modifying~\cite{loquercio2018dronet} to reduce the number of parameters in the neural network, while enhancing connectivity in the fully-connected layer. The decoder consists of deconvolutional layers stacked with activation functions without skip connections. The network architecture used in this work is depicted in Figure~\ref{fig:VAE_network_architecture}, where its fine tuning was guided by the observation that increasing the size of the fully connected hidden layers tended to enhance the quality of reconstruction from the network. The loss term is defined as:

\small
\begin{equation}
    \mathcal{L} = \mathcal{L}_{recon} + \beta_{norm}\mathcal{L}_{KL},
\end{equation}
\normalsize
where $\mathcal{L}_{KL}$ denotes the KL-divergence loss defined as:

\small
\begin{equation}
    \mathcal{L}_{KL}(\bm{\mu}, \bm{\sigma}) = \frac{1}{2}\sum_{j=1}^J \left (1 + \log((\bm{\sigma}_j)^2) - (\bm{\mu}_j)^2 - (\bm{\sigma}_j)^2 \right).
\end{equation}
\normalsize
The constant $\beta_{norm} = \frac{\beta\cdot J}{H\cdot W}$ weighing the KL-divergence loss is normalized accounting for the latent space and image dimensions. The value of $\beta$ (here $\beta = 1$) allows the mixing of the contributions from the KL-divergence loss and the reconstruction loss and is a tunable hyperparameter~\cite{higgins2017betavae}. The depth cameras used to collect the datasets consist of a stereo pair and are unable to reconstruct depth from textureless features and miss the depth data around the edges of features resulting in a ``stereo shadow'' region. We design a semantic-weighted reconstruction loss function $\mathcal{L}_{recon}$ for training the neural network taking into consideration these errors from such sensors by ignoring the contribution of the invalid pixels in the loss function. The semantic-weighted reconstruction loss is defined as:
\vspace{-1ex}
\begin{multline}\label{eq:reconstruction_loss}
    \mathcal{L}_{recon}(\mathbf{x}, \mathbf{x}_{recon}, \mathbf{x}_{val}, \mathbf{x}_{seg}) =\\ \sum_{h=1}^{H}\sum_{w=1}^{W}((\mathbf{x}-\mathbf{x}_{recon})^2\odot{\mathbf{x}_{val}}\odot{\lambda(\mathbf{x}_{seg})})_{(h,w)},
\end{multline}
where $\mathbf{x}_{recon}$ is the reconstructed image and $\{\mathbf{x}, \mathbf{x}_{val}, \mathbf{x}_{seg}\}$ represents one sample from the dataset $\mathbf{X}$. The use of $\mathbf{x}_{val}$ during \ac{sevae} training eliminates the contribution of the invalid input pixels from the loss function, allowing the network to represent distributions consisting of information in the valid pixels. The function $\lambda(\mathbf{x}_{seg})$ creates a pixel-wise weight mask to give higher weight to the pixels that correspond to disproportionately thin obstacles using the semantic label information $\mathbf{x}_{seg}$ (if available). The weight $\nu_\ell$ of a pixel $\ell$ belonging to a semantic instance $S_k \in \mathbb{S}$, where $\mathbb{S}$ is the set of all instances of semantics in $\mathbf{x}_{seg}$, depends on the pixel count $p_k$ of instance $S_k$. The weight of the pixels not belonging to a semantic instance is set to 1. Formally, the weight $\nu_\ell$ is defined as:

\small
\begin{equation}
\nu_\ell = 
    \begin{cases}
    \max(W_{const}/p_k, \nu_{\min}),~ \ell \in S_k~\textrm{and}~p_k > p_{\min}\\
    1,~\textrm{otherwise}
    \end{cases},
\end{equation}
\normalsize
where the term $W_{const}$ (here $W_{const}=6000$) acts as a multiplicative constant to weigh the inverse count of pixels per semantic, while $\nu_{\min}$ (here $\nu_{\min}=15$) limits the minimum per-pixel weight. This weighing term is applied to magnify the contribution of small-sized semantics while allowing larger-sized semantics to be proportionally weighed based on the number of pixels. We ignore semantics smaller than $p_{\min}$ (here $p_{\min}=40$ pixels) to prevent the \ac{sevae} from trying to reconstruct extremely small regions, that may, in the real-world correspond to sensor noise and data imperfections. Given the loss $\mathcal{L}$ and the semantic information, the \ac{sevae} is trained to encode a depth image into the low-dimensional latent representation $\boldsymbol{\mu}$ to be used for collision prediction.

\subsection{Collision Prediction Network and Action Planning}\label{sec:collision_prediction_network}

\begin{figure}[h]
\includegraphics[width=\columnwidth]{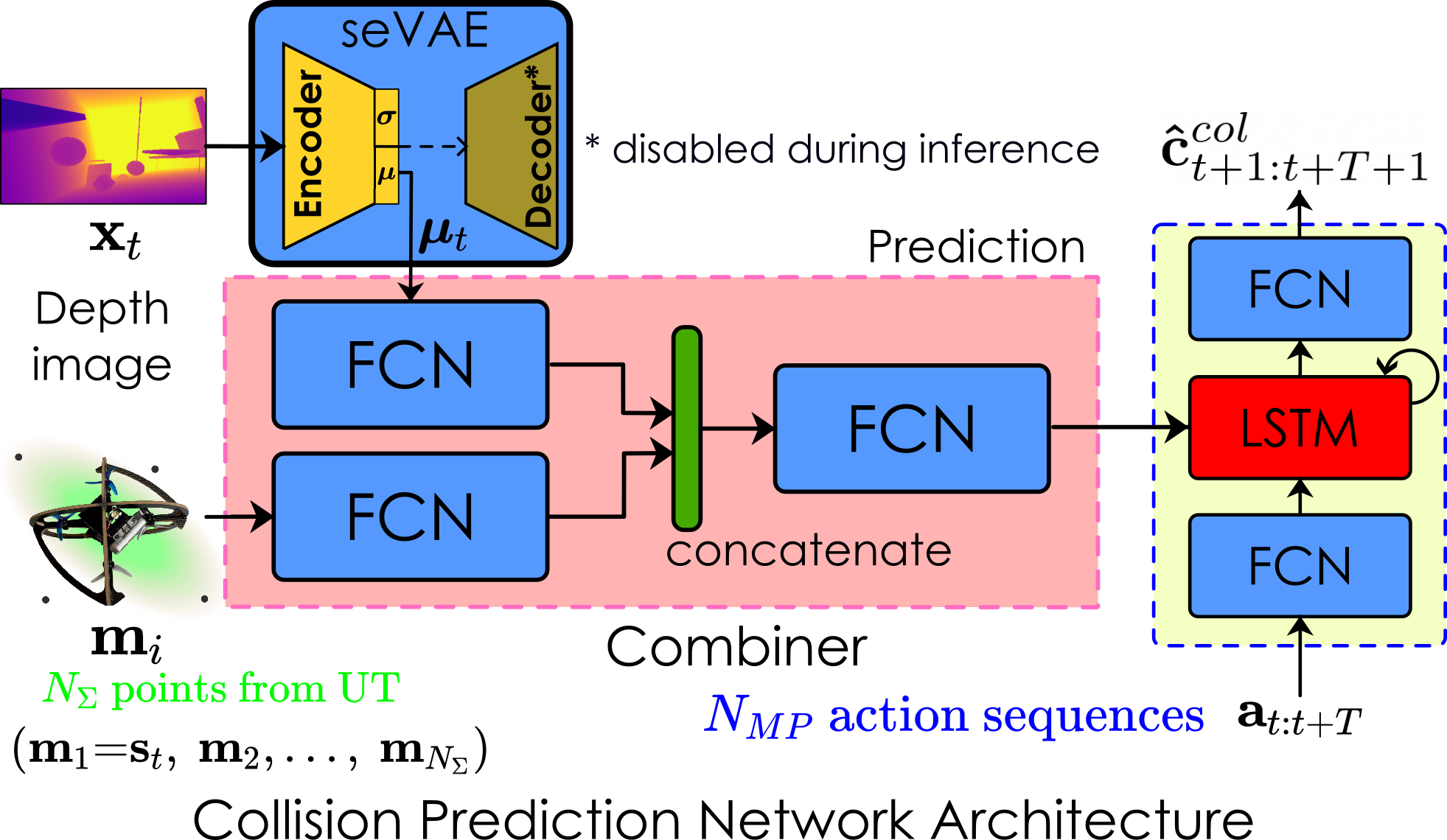}
\vspace{-4ex}
\caption{Proposed architecture for the modularized approach involving the \ac{sevae} and the Collision Prediction Network (CPN). The \ac{sevae} encodes the input depth image $\mathbf{x}_t$ into the latent representation $\bm{\mu}_t$ which is used by the \ac{cpn} to predict the collision scores $\hat{\mathbf{c}}^{col}_{t+1:t+T+1}$ for each action sequence $\mathbf{a}_{t:t+T}$ in the motion primitives library. Furthermore, the method utilizes the $N_{\Sigma}$ sigma points calculated based on $\mathbf{s}_t$ and $\bm{\Sigma}_t$ to calculate the robot's partial state uncertainty-aware collision score $\hat{c}^{uac}$.}
\vspace{-2ex}
\label{fig:CPN_network_architecture}
\end{figure}

The latent space from the \ac{sevae} is then exploited by the \ac{cpn} to facilitate safe navigation building upon our previous work ORACLE~\cite{ORACLE_ICRA2022} and extending it to $3$D navigation, while retaining the benefits of the modularized \ac{sevae}-based latent space including its ability to assimilate real sensor data. To this end, the CNN part of the original \ac{cpn} in ORACLE is replaced by the \ac{sevae} which takes the current depth image $\mathbf{x}_t$ and outputs the compressed latent vector $\boldsymbol{\mu}_t$, that is then fed to the \ac{cpn} as illustrated in Figure~\ref{fig:CPN_network_architecture}.

Specifically, let $\Bs,\Vs$ be the body frame and vehicle frame of the robot respectively, and $\mathbf{s}_t = [{\mathbf{v}_t}^T, \omega_t, \vartheta_t, \varphi_t]^T $ the estimated partial state of the robot at time $t$ consisting of a) the $3$D velocity in $\Vs$ ($\mathbf{v}_t=[v_{t,x}, v_{t,y}, v_{t,z}]^T \in \mathbb{R}^{3 \times 1}$), b) the angular velocity around the $z$-axis of the body-frame $\Bs$ ($\omega_t$), as well as c) the roll ($\vartheta_t$) and pitch angles ($\varphi_t$). Let $\boldsymbol{\Sigma}_t$ denote the covariance of the estimated robot’s partial state, $\mathbf{n}^g_t$ the $3$D unit goal vector, expressed in $\Vs$, that the robot has to follow, $\psi_t$ the current yaw angle of the robot, and $\mathbf{a}_{t:t+T}=[\mathbf{a}_t, \mathbf{a}_{t+1},\hdots,\mathbf{a}_{t+T-1}]$ an action sequence having length $T$ where the action at time step $t+i~(i=0,\hdots,T-1)$ includes a) the reference $3$D velocity expressed in the vehicle frame $\mathbf{v}^r_{t+i}$ and
b) the steering angle ($\delta^r_{t+i}$) from the current yaw angle of the robot ($\psi_{t}$), such that $\mathbf{a}_{t+i}=[{\mathbf{v}^r_{t+i}}^T, \delta^r_{t+i}]^T$.
The method finds an optimized collision-free sequence of actions $\mathbf{a}_{t:t+T}$ given $(\mathbf{x}_t,\mathbf{s}_t,\boldsymbol{\Sigma}_t)$ (specifically using the latent vector $\bm{\mu}_t$ derived from $\mathbf{x}_t$ using the \ac{sevae}'s encoder) enabling the robot to safely navigate to the goal vector $\mathbf{n}^g_t$. 

The \ac{cpn} processes $\bm{\mu}_t$, $\mathbf{s}_t$, and a Motion Primitives-based Library (MPL) of $N_{MP}$ sequences of future velocity and steering angle references $\mathbf{a}_{t:t+T}$ to predict the collision scores $\hat{\mathbf{c}}^{col}_{t+1:t+T+1}=[\hat{c}^{col}_{t+1}, \hat{c}^{col}_{t+2},\hdots,\hat{c}^{col}_{t+T}]$ of the robot at each time step from $t+1$ to $t+T$ in the future for each action sequence. 
To account for the uncertainty in the robot's partial state, we first calculate $N_{\Sigma} = 2\gamma + 1$ sigma points ($\mathbf{m}_1=\mathbf{s}_t$, $\hdots$, $\mathbf{m}_{N_{\Sigma}}$), where $\gamma$ is the dimension of the robot's partial state (here $\gamma=6$), based on the mean $\mathbf{s}_t$ and covariance $\bm{\Sigma}_t$ using the Unscented Transform (UT) \cite{Julier1997unscented}. Using the sigma points, the uncertainty-aware collision score $\hat{c}^{uac}$ is calculated as presented in~\cite{ORACLE_ICRA2022}. A set of safe action sequences is derived by thresholding $\hat{c}^{uac}$ and the action sequence that leads to the end velocity of the robot best aligned with $\mathbf{n}^g_t$ is chosen. The first action in the sequence is executed, while the process is repeated in a receding horizon fashion. Notably, in ORACLE, the \ac{cpn} is trained end-to-end entirely in simulation and requires a computationally-expensive pre-processing step in the depth image input to close the sim-to-real gap. In this work, the \ac{cpn} is trained with the frozen weights of the \ac{vae} in simulation environments which have large obstacles with randomized shapes and thin obstacles with diameters less than $5~\textrm{cm}$, as illustrated in Figure~\ref{fig:data_collection_diagram}.3.

\section{Evaluation Studies}\label{sec:evaluation}
The proposed method, and its submodules, were extensively evaluated as presented below. 

\subsection{Training Methodology}\label{sec:training_dataset}
A composite dataset of both real and simulated images was collected for training the VAE network. Real data was collected with an Intel RealSense D$455$ depth camera. This dataset consists of images in confined spaces, indoor rooms, long corridors, and outdoor environments with trees. Thin obstacles such as tree branches, rods, and poles are manually labeled in this dataset. We also utilize images from the NYU Depth Dataset v2~\cite{silberman2012nyu_depth}. Simulation images are collected using both Gazebo Classic~\cite{gazbeoclassic} and Isaac Gym Simulators~\cite{isaac_gym}. Isaac Gym offers a segmentation camera providing instance segmentation masks for the corresponding depth images. Simulated meshes having cross sections below a size of $5~\textrm{cm}$ are assigned semantic instance IDs. The segmentation camera allows rendering images with pixel values equal to the ID of the semantic occupying that pixel. The aggregated dataset $\mathbf{X}$ consists of $\sim 66,000$ images, of which $\sim 35,000$ are simulated. $21,000$ simulated images have semantic labels, and $\sim 1,000$ real images are labeled. The training and validation sets are split with a $80\%:20\%$ ratio from this dataset. We train the network using the Adam Optimizer~\cite{kingma2015adam} with a learning rate of $10^{-4}$ for $40~\textrm{epochs}$.

\begin{figure}[h]
\includegraphics[width=\columnwidth]{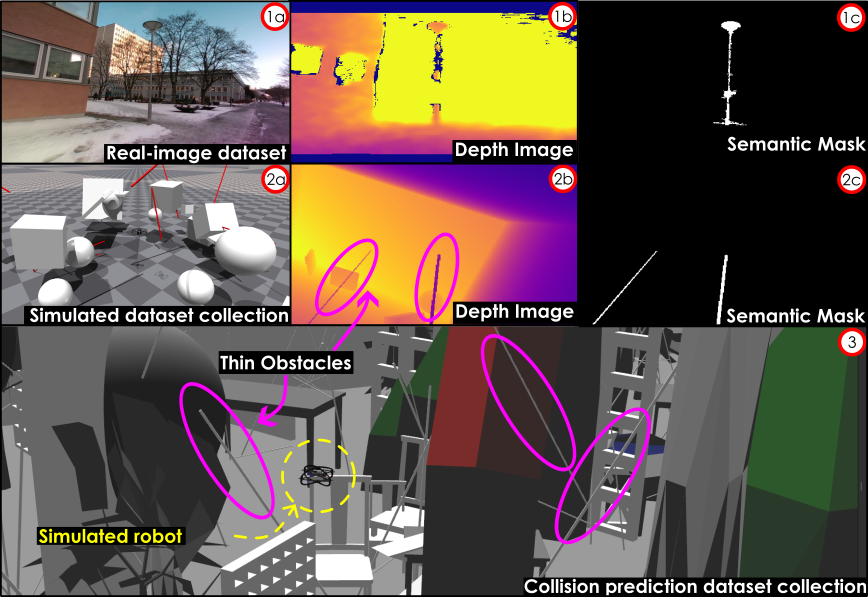}
\vspace{-4ex}
\caption{Instances of depth images (1b, 2b) collected for the \ac{sevae} training from the real world (1a) and the IsaacGym (2a) simulator are shown. RGB images (1a) from a collected dataset (if available) are used to create semantic labels (1c) for the aligned depth images. The objects beyond the sensing range of the depth camera are not labeled. The IsaacGym simulator provides semantic labels for thin obstacles (2c). Simulation data for training the \ac{cpn} is collected in Gazebo Classic (3) with a simulated drone in an obstacle-filled environment. Randomized action sequences are executed to collect a collision dataset.}
\vspace{-1ex}
\label{fig:data_collection_diagram}
\end{figure}

Collision datasets for training the \ac{cpn} are collected using the Gazebo-based RotorS simulator~\cite{Furrer2016rotors} with a simulated robot model. An environment consisting of obstacles having different shapes and sizes is constructed and thin obstacles with cross sections smaller than $5~\textrm{cm}$ are introduced as shown in Figure~\ref{fig:data_collection_diagram}. Randomized action sequences $\mathbf{a}_{t:t+T}$ within the robot's FOV are generated and executed until a collision or a timeout occurs. Datapoints consisting of $d = (\mathbf{x}_t, \mathbf{s}_t, \mathbf{a}_{t:t+T}, \mathbf{\hat{c}}^{col}_{t+1:t+T+1})$ are collected. The data is also augmented by performing a horizontal flip and appending random actions at the end of the collision episodes, similar to~\cite{ORACLE_ICRA2022}. The depth images $\mathbf{x}_t$ from this are passed through the learned \ac{sevae} model to obtain its latent representation $\bm{\mu}_t$. Finally, datapoints  $d^\prime = (\bm{\mu}_t, \mathbf{s}_t, \mathbf{a}_{t:t+T}, \mathbf{\hat{c}}^{col}_{t+1:t+T+1})$ are created replacing the depth image $\mathbf{x}_t$ with its corresponding latent representation $\bm{\mu}_t$ to train the \ac{cpn}.

\begin{figure}[h]
\includegraphics[width=\columnwidth]{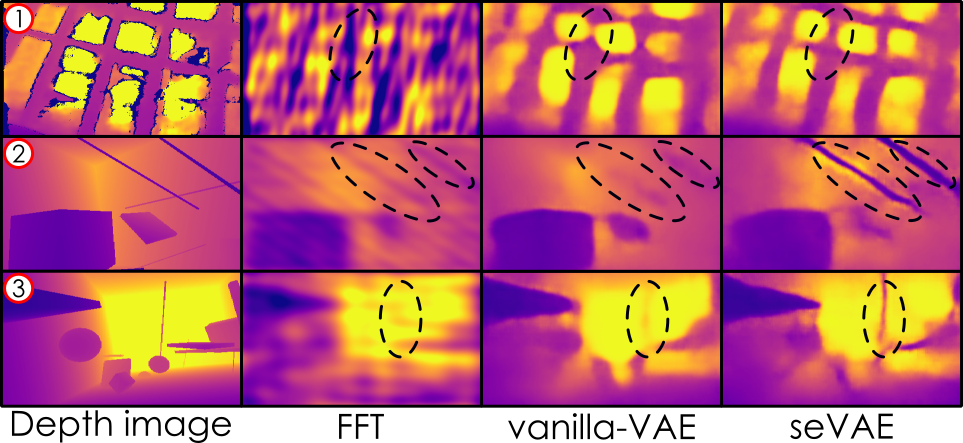}
\vspace{-4ex}
\caption{The proposed method is compared with Fast Fourier Transform (FFT) and vanilla-VAE without semantic weighted loss on real (1) and simulated images (2,3). The FFT reconstruction is created from the largest $64$ magnitudes in the frequency domain representation, while the vanilla-VAE and the semantically-enhanced VAE use a $128$ dimensional latent vector. Ellipsoids are drawn around expected reconstructions of thin obstacles from the input depth images.}
\label{fig:vae_fft_reconstruction_comparison}
\vspace{-4ex}
\end{figure}

\subsection{Comparison of reconstruction methods for thin obstacles}\label{sec:reconstruction_comparison}
First, the proposed semantically-enhanced VAE is evaluated by comparing it against a baseline compression based on the Fast Fourier Transform (FFT) and a vanilla-VAE trained without the weighing function $\lambda(\mathbf{x}_{seg})$. To allow for a fair comparison with the $128$ dimensional latent variable $\bm{\mu}$ of the VAEs, the FFT reconstructions are generated from the $64$ complex frequency-domain representations having the largest magnitudes while zeroing out the others. Reconstructions from the compared methods are depicted in Figure~\ref{fig:vae_fft_reconstruction_comparison}. 
The FFT reconstruction is unable to preserve information at this resolution especially for the features corresponding to smaller-sized obstacles. Additionally, the presence of sensor noise in real images degrades the performance of the FFT even further. The vanilla-VAE is able to reconstruct larger parts of the images well but misses out on the regions with small cross-section (thin) obstacles for both real and simulated images. The \ac{sevae} is able to reconstruct regions with smaller cross sections better than the above approaches. The key role of semantically-augmented training is particularly visible in Figure~\ref{fig:vae_fft_reconstruction_comparison}. We statistically compare the performance of the three approaches to consider the MSE over the whole image and also specifically over only the semantically labeled pixels (thin obstacles) of an image and present the results in Table~\ref{tab:mse_comparison_table_fft_vanilla_vae}. For comparison, we normalize the valid pixel values between $0$ and $1$. As shown, \ac{sevae} presents a significantly better performance in reconstructing the semantic regions, with a relatively small reduction in the performance over the whole image against the vanilla-VAE using the same number of parameters, while also outperforming FFT-based compression. It is noted that relatively inferior reconstruction quality is less significant for large objects, as long as collision-avoidance is concerned, but it is critical not to miss thin obstacles.

\begin{table}[h]
    \vspace{-2ex}
    \centering
    \caption{Comparison of MSE for reconstructed images with different methods (seVAE, vanilla-VAE, FFT-based).}
    \vspace{-2ex}
    \begin{tabularx}{\columnwidth}{|l|X|X|X|}
      \hline
      \multicolumn{4}{|c|}{\textbf{Simulated Images (Count: $2239$)}} \\
      \hline
      \textbf{MSE over:} & \textbf{FFT} & \textbf{vanilla-VAE} & \textbf{seVAE}\\
       \hline
      Entire image & $553.21$ & $\mathbf{276.64}$ & $404.50$  \\
      \hline
      Semantic pixels  & $82.14$  & $89.51$  & $\mathbf{22.13}$  \\
      \hline
      \multicolumn{4}{|c|}{\textbf{Real Images (Count: $365$)}} \\
      \hline
      \textbf{MSE over:} & \textbf{FFT} & \textbf{vanilla-VAE} & \textbf{seVAE}\\
      \hline
      Entire image & $2804.28$ & $\mathbf{350.01}$ & $420.35$  \\
      \hline
      Semantic pixels  & $156.95$  & $69.16$  & $\mathbf{28.41}$  \\
      \hline
    \end{tabularx}
    \label{tab:mse_comparison_table_fft_vanilla_vae}
    \vspace{-4ex}
\end{table}

\begin{figure}[h!]
\centering
\includegraphics[width=0.9\columnwidth]{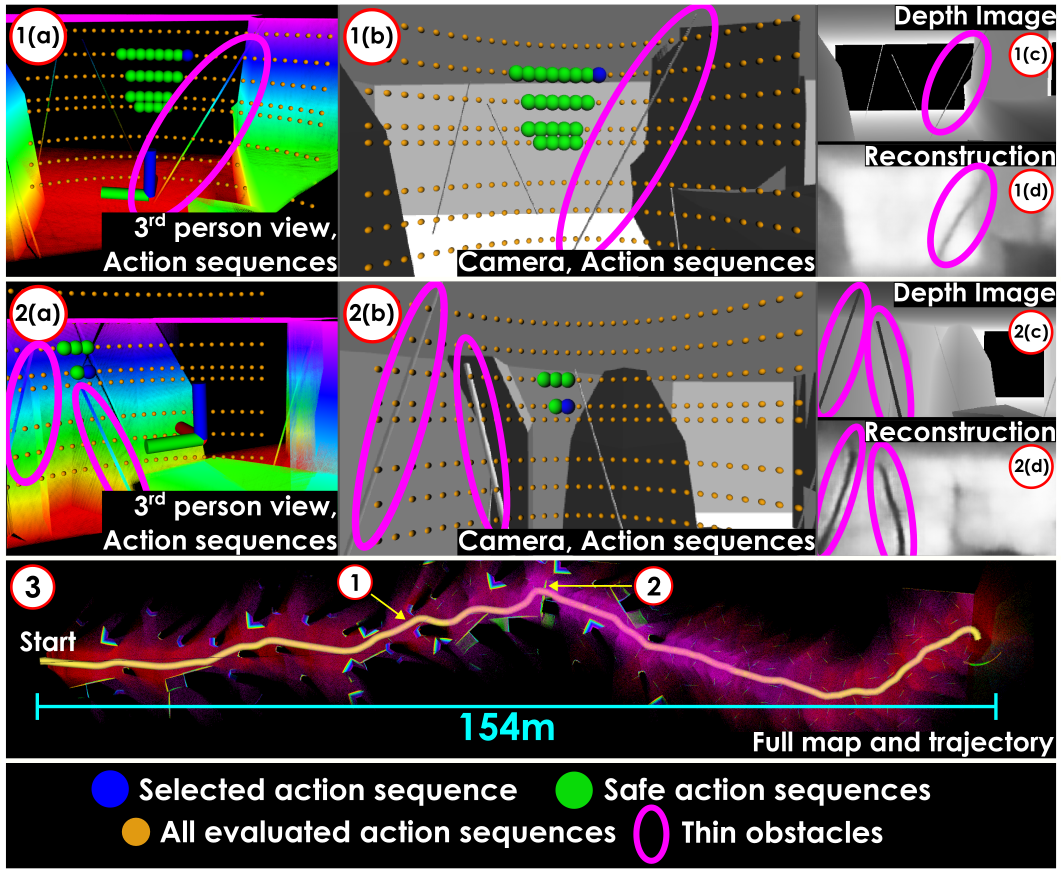}
\caption{Results from one of the simulation missions conducted in a Gazebo environment with diverse obstacle configurations including thin obstacles (cross section $4~\textrm{cm}$). The sub-figures 1 and 2 show two instances during the mission detailing a) the point cloud, b) the onboard camera image both overlaid with markers representing estimated endpoints of collision-free action sequences, c) the raw depth image, and d) reconstructed depth image from \ac{sevae}. Sub-figure 3 shows the final map and the trajectory followed by the robot. The simulation shows the ability of the \ac{sevae} to reconstruct thin and regular objects well, and the overall method to avoid obstacles.}
\vspace{-4ex}
\label{fig:vae_sim}
\end{figure}

\subsection{Simulation Studies}\label{sec:sim_studies}
Subsequently, the proposed method is evaluated in complex simulation studies involving diverse obstacle configurations including distributed thin objects such as rods. We design three simulation worlds consisting of three sections each, with different kinds of obstacles placed using Poisson disc sampling. The first section contains exclusively large-sized obstacles with Poisson disc sampling radius $r_1$, the second section contains both large sized, and thin obstacles with cross section $4~\textrm{cm}$, sampled independently of each other with radii $r_2$ and $r_3$ respectively, and the third section consists only of thin rods with a similar cross section sampled with radius $r_4$. Each section is $50~\textrm{m}\times50~\textrm{m}$ and placed serially. The robot is commanded to travel $150~\textrm{m}$ along the course at a speed of $1.0~\textrm{m/s}$. To derive the UT sigma points, we use $\boldsymbol{\Sigma}_t = \textrm{diag}({\sigma_v}^2, {\sigma_v}^2, {\sigma_v}^2, 0, 0, 0)$, with ${\sigma_v}=0.2~\textrm{m}/\textrm{s}$. We perform $20$ runs of the experiment per environment for both the proposed method (trained with datapoints $d^\prime$) and a $3$D extension of our end-to-end ORACLE method~\cite{ORACLE_ICRA2022} (trained with datapoints $d$) with different initial positions and orientations. A few instances and the full path from one of the missions of the proposed method are shown in Figure~\ref{fig:vae_sim} and the statistical results are logged in Table~\ref{tab:sim_result}. The environment sampling variables are listed beside each environment in the format ($r_1$, $r_2$, $r_3$, $r_4$). It is noted that the ensemble of neural networks is not used for ORACLE to have a fair comparison. 
The proposed method outperforms ORACLE with a higher success rate in terms of completing the course without collision, in denser environments.

\begin{table}[h]
 \vspace{-2ex}
 \caption{Comparison between success rates of end-to-end vs proposed method in simulation.}
 \vspace{-2ex}
\centering
\begin{tabular}{|c|c|c|}  
\hline
\textbf{Environment ($r_1$, $r_2$, $r_3$, $r_4$)} & \textbf{Method} & \textbf{Success \%} \\
\hline
    \multirow{2}{*}{Sparse ($6.5$, $6.5$, $4.5$, $3.5$)}
       & Proposed method & $95\%$  \\
        & End-to-end ORACLE        & $95\%$     \\
\hline
     \multirow{2}{*}{Medium ($6.25$, $6.25$, $3.5$, $3.0$)} & Proposed method          & $\mathbf{80}\%$ \\
      & End-to-end ORACLE     & $75\%$     \\            
\hline
     \multirow{2}{*}{Dense ($6.0$, $6.0$, $3.0$, $2.5$)} & Proposed method          & $\mathbf{60}\%$   \\
       & End-to-end ORACLE     & $45\%$  \\
\hline
\end{tabular}
\label{tab:sim_result}
 \vspace{-4ex}
\end{table}

\begin{figure*}[ht!]
\includegraphics[width=\textwidth]{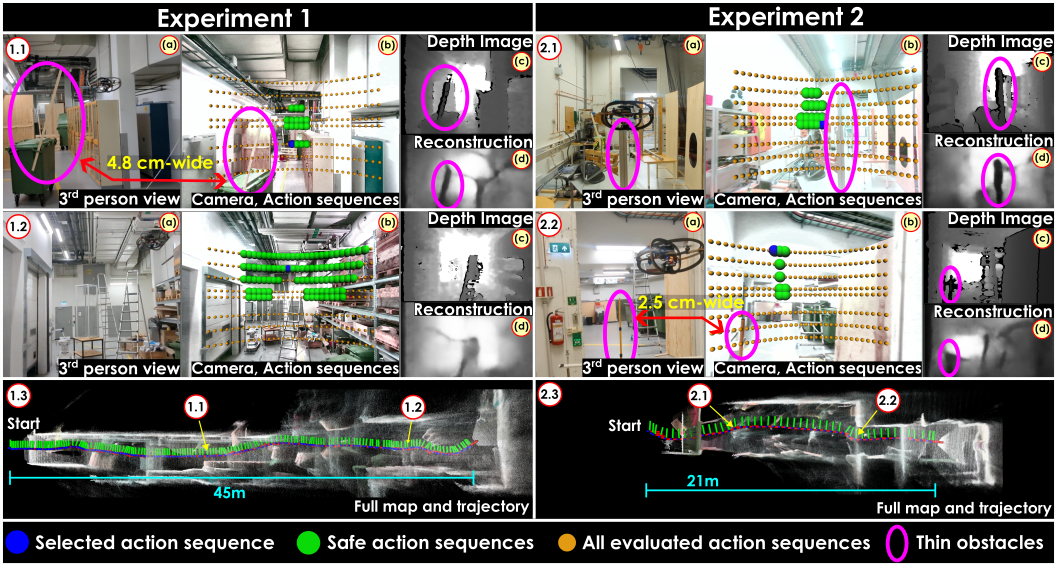}
\vspace{-4ex}
\caption{Results from the two real-world experiments conducted to evaluate the proposed method. The first took place in a long cluttered corridor ($\sim 50~\textrm{m}$) including thin obstacles as small as $4.8~\textrm{cm}$ in width. The second experiment was conducted in a relatively smaller corridor ($\sim 20~\textrm{m}$) with predominantly thin obstacles, some as small as $2.5~\textrm{cm}$ in width, some with texture-less reflective surfaces. The sub-figures 1.1 and 1.2 show two instances during experiment 1 (similarly 2.1, 2.2 for experiment 2) detailing a) $3^{\textrm{rd}}$ person view of the robot and the environment, b) onboard color camera image overlaid with markers representing estimated endpoints of collision-free action sequences, c) raw depth image, and d) reconstructed depth image from \ac{sevae}. 1.3 (similarly 2.3 for experiment 2) shows the final map and trajectory followed by the robot in experiment 1. The reconstructed maps are only for visualization purposes and are not computed onboard the robot during flights. It can be seen that the method is successfully able to respond to and avoid all the obstacles.}
\vspace{-2ex}
\label{fig:experiment_diagram}
\end{figure*}

\subsection{Experimental Evaluation}\label{sec:experimental_evaluation}

Beyond simulations, the proposed method was evaluated in two real-world confined environments further involving thin obstacles using a collision-tolerant aerial robot design similar to \cite{rmfowl_icuas} which integrates a sensor suite including an Intel Realsense D455 RGB-D sensor, an Intel Realsense T265 visual-inertial module, an mRo Pixracer flight controller, and an NVIDIA Xavier NX board. To derive the UT sigma points, we use $\boldsymbol{\Sigma}_t = \textrm{diag}({\sigma_v}^2, {\sigma_v}^2, {\sigma_v}^2, 0, 0, 0)$, with ${\sigma_v}=0.2~\textrm{m}/\textrm{s}$. One experiment is performed (with an average robot speed of $1.0~\textrm{m/s}$) in a long cluttered corridor with thin obstacles (as small as $4.8~\textrm{cm}$ in width) obstructing the path of the robot. The goal point is defined $50~\textrm{m}$ from the starting point along the corridor and expressed as a goal direction for the navigation method exploiting only a partial state without position information. A second experiment is conducted (with an average robot speed of $0.75~\textrm{m/s}$) in a smaller environment with challenging hard-to-perceive obstacles with small cross-sections (as small as $2.5~\textrm{cm}$ in width), and texture-less reflective surfaces. The size and the material of the obstacles in these experiments pose challenges for accurate depth reconstruction using the depth camera onboard the robot. The resolution of the RealSense D455 is set to $480\times640~\textrm{pixels}$ and the depth image is downsampled to the input resolution of the encoder ($270\times480~\textrm{pixels}$) before encoding. The inference times of \ac{sevae} and \ac{cpn} on the platform are $11~\textrm{ms}$ and $29~\textrm{ms}$ respectively. The results of these experiments are shown in Figure~\ref{fig:experiment_diagram}.

\begin{figure}[ht!]
\centering
\includegraphics[width=0.9\columnwidth]{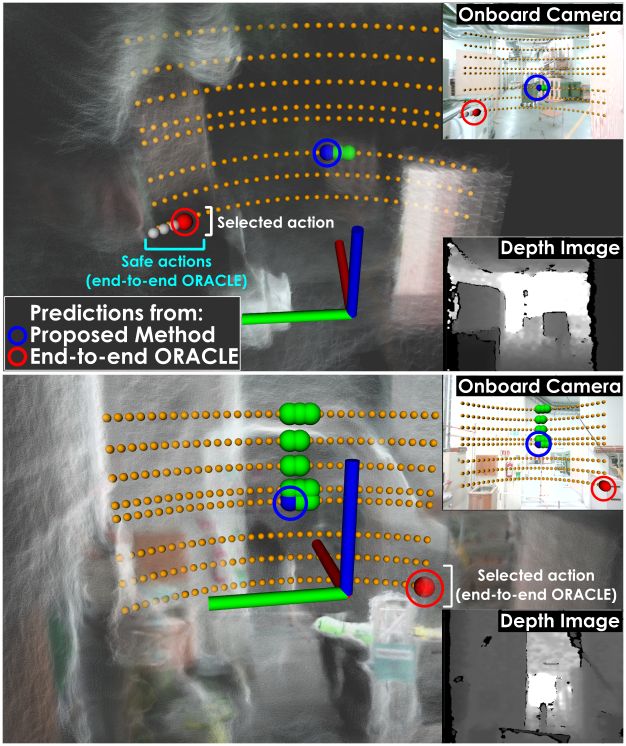}
\caption{Comparison of collision predictions of ORACLE end-to-end trained purely in simulation against the proposed method with un-filtered depth data in the real-world experiments. Our proposed method predicts collision-free action sequences and selects the safest one toward the goal whereas ORACLE incorrectly predicts the collision-free action sequences.}
\vspace{-3ex}
\label{fig:vae_vs_end_to_end_oracle}
\end{figure}

Finally, the proposed approach is also experimentally compared against the end-to-end trained ORACLE as in Section~\ref{sec:sim_studies}. Specifically, the collision prediction step of ORACLE was ran for the depth frames of the above experiments. Notably, originally ORACLE also runs a filter that attempts to fill-in invalid pixels coming from the sensor~\cite{ku2018defense}, which is however computationally burdensome ($\sim20~\textrm{ms}$ compute time on an NVIDIA Xavier NX) and not needed in the proposed method thus not executed for the purposes of fair comparison. As shown in Figure~\ref{fig:vae_vs_end_to_end_oracle}, the end-to-end ORACLE incorrectly predicts action sequences --that may cause the robot to collide with the environment-- as collision-free, unlike the proposed approach which enables safe action sequences towards the goal direction. Despite the potential benefits of end-to-end training, ORACLE is unable to incorporate the perceptual focus on thin obstacles of the new exploit using the \ac{sevae} and is also unable to exploit real sensor observations in the training. As a result, learning to avoid hard-to-perceive thin obstacles is only trained from simulated data which present significant differences compared to real data (not merely white noise and other easy-to-simulate errors)~\cite{yang2018segstereo,ahn2019analysis,mallick2014characterizations} thus widening the sim-to-real gap. Note that, the ensemble of neural networks in~\cite{ORACLE_ICRA2022} is not used to have a fair comparison.  


\section{Conclusions}\label{sec:conclusions}
This paper presented a modularized learning-based navigation method that can navigate cluttered environments involving hard-to-perceive obstacles, without access to a map or the full robot state. A semantically-enhanced VAE is designed that can encode raw sensor observations into a latent representation. This is trained with both simulated and real images and utilizes semantic labels to better encode the hard-to-perceive thin obstacles. A Collision Prediction Network is then trained completely in simulation with the latent space from the \ac{sevae} to learn to predict collision scores for a set of action sequences. To evaluate our approach, first we compare the performance of the \ac{sevae} with other methods for compression. Then, simulation studies are performed to compare the collision avoidance performance of our method with an end-to-end trained (in simulation) approach. Furthermore, we conduct real experiments to show the performance of our method in cluttered environments containing hard-to-perceive thin obstacles, reflective and textureless surfaces. Finally we highlight the benefits of this method in handling real sensor data errors compared to methods trained only simulation. The method and datasets used in this work will be open-sourced to the community.



\bibliographystyle{IEEEtran}
\bibliography{0_main}

\end{document}